\documentclass[letterpaper, 10 pt, journal, twoside]{ieeetran}  

\usepackage{amsmath}
\usepackage{amssymb}  
\usepackage{booktabs}
\usepackage{graphicx}
\usepackage{cite}
\usepackage{xcolor}
\usepackage{scalerel}
\usepackage{subfig}
\usepackage{mathtools}

\newcommand{\vect}[1]{\mathbf{ #1}}

\newcommand{\vectg}[1]{{\boldsymbol{ #1}}}

\newcommand{\T}{^\mathsf{T}}

\newcommand{\argmin}{\operatornamewithlimits{argmin}}

\newcommand{\vA}{\vect{A}}

\newcommand{\vC}{\vect{C}}
\newcommand{\vD}{\vect{D}}

\newcommand{\vK}{\vect{K}}

\newcommand{\vP}{\vect{P}}
\newcommand{\vQ}{\vect{Q}}
\newcommand{\vR}{\vect{R}}

\newcommand{\vU}{\vect{U}}
\newcommand{\vV}{\vect{V}}

\newcommand{\vZ}{\vect{Z}}

\newcommand{\vb}{\vect{b}}

\newcommand{\vl}{\vect{l}}

\newcommand{\vp}{\vect{p}}
\newcommand{\vq}{\vect{q}}

\newcommand{\vs}{\vect{s}}
\newcommand{\vt}{\vect{t}}
\newcommand{\vu}{\vect{u}}

\newcommand{\vw}{\vect{w}}
\newcommand{\vx}{\vect{x}}

\newcommand{\vbeta}{\vectg{\beta}}

\newcommand{\vpi}{\vectg{\pi}}

\newcommand{\vLambda}{\vectg{\Lambda}}

\newcommand{\vtheta}{\vectg{\theta}}

\newcommand{\vSigma}{\vectg{\Sigma}}

\newcommand{\cN}{\mathcal{N}}

\newcommand{\cP}{\mathcal{P}}

\title{QuadricSLAM: Dual Quadrics from Object Detections as Landmarks in Object-oriented SLAM}

\author{Lachlan Nicholson, Michael Milford, and Niko S\"underhauf%
\thanks{Manuscript received April 8, 2018; Revised June 30, 2018; Accepted July 29, 2018. This paper was recommended for publication by Editor C. Stachniss upon evaluation of the Associate Editor and Reviewers' comments. This research was conducted by the Australian Research Council Centre of Excellence for Robotic Vision (project number CE140100016).}

\thanks{The authors are with the ARC Centre of Excellence for Robotic Vision, Queensland University of Technology (QUT), Brisbane, Australia (e-mail: lachlan.nicholson@hdr.qut.edu.au; michael.milford@qut.edu.au; niko.suenderhauf@qut.edu.au). Michael Milford is supported by an Australian Research Council Future Fellowship (FT140101229).}

\thanks{The authors gratefully thank John Skinner for his contributions to the evaluation environment, and Madeline Miller for her help annotating data.}

\thanks{Digital Object Identifier (DOI): see top of this page.}
}

\begin{document}

\maketitle
\markboth{IEEE Robotics and Automation Letters. Preprint Version. Accepted July, 2018} 
{Nicholson \MakeLowercase{\textit{et al.}}: QuadricSLAM}

\begin{abstract}
    In this paper, we use 2D object detections from multiple views to simultaneously estimate a 3D quadric surface for each object and localize the camera position. 
    We derive a SLAM formulation that uses dual quadrics as 3D landmark representations, exploiting their ability to compactly represent the size, position and orientation of an object, and show how 2D object detections can directly constrain the quadric parameters via a novel geometric error formulation. We develop a sensor model for object detectors that addresses the challenge of partially visible objects, and demonstrate how to jointly estimate the camera pose and constrained dual quadric parameters in factor graph based SLAM with a general perspective camera.
\end{abstract}


\section{Introduction}

\IEEEPARstart{R}{ecently}, the performance of vision-based object detection has seen impressive improvements resulting from the ``rebirth'' of Convolutional Neural Networks (ConvNets). Building on the groundbreaking work by Krizhevsky et al.~\cite{Krizhevsky12} and earlier work~\cite{LeCun98, LeCun04}, several other groups (e.g. \cite{Girshick14, Szegedy15, Ren15, Liu16, He17}) have increased the quality of ConvNet-based methods for object detection. Recent approaches have even reached human performance on the standardized ImageNet ILSVRC benchmark \cite{Russakovsky15} and continue to push the performance boundaries on other benchmarks such as COCO~\cite{Lin14}.

Despite these impressive developments, the Simultaneous Localization And Mapping community (SLAM) has not yet fully adopted the newly arisen opportunities to create semantically meaningful maps. SLAM maps typically represent \emph{geometric} information, but do not carry immediate object-level \emph{semantic} information. Semantically-enriched SLAM systems are appealing because they increase the richness with which a robot can understand the world around it, and consequently the range and sophistication of interactions that a robot may have with the world, a critical requirement for their eventual widespread deployment at work and in homes.

\begin{figure}[t]
    \centering
    \includegraphics[width=0.49\linewidth]{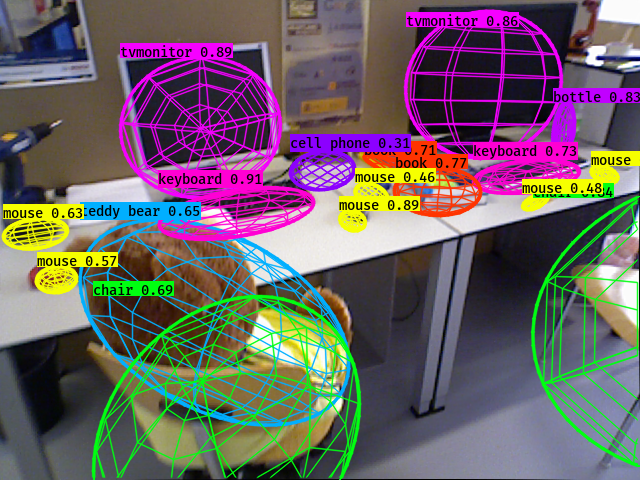}
    \includegraphics[width=0.49\linewidth]{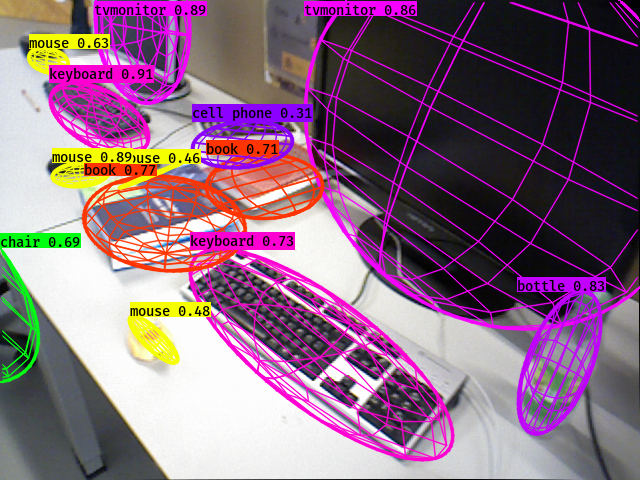}
    \caption{QuadricSLAM uses \emph{objects} as landmarks and represents them as constrained dual quadrics in 3D space. 
    QuadricSLAM jointly estimates camera poses and quadric parameters from odometry measurements and object detections, implicitly performing loop closures based on the object observations. 
    This figure illustrates how well the estimated quadrics fit the true objects when projected into the camera images from different viewpoints.}
    \label{fig:intro}
\end{figure}

Semantically meaningful maps should be object-oriented, with objects as the central entities of the map. \emph{Quadrics}, i.e. 3D surfaces such as ellipsoids, have a number of attractive properties as landmark representations for object-oriented semantic maps (see Figure~\ref{fig:intro}). 
In contrast to more complex object representations such as truncated signed distance fields~\cite{Curless96}, quadrics have a very compact representation and can be manipulated efficiently within the framework of projective geometry. Quadrics capture information about the size, position, and orientation of an object, and can serve as anchors for more detailed 3D reconstructions if necessary.
They are also appealing from an integration perspective: as we are going to show, in their \emph{dual} form, quadrics can be constructed directly from object detection bounding boxes and conveniently incorporated into a factor graph based SLAM formulation.

In this paper we make the following contributions. We first show how to parametrize object landmarks in SLAM as \emph{constrained dual quadrics}. We then demonstrate that visual object detection systems such as YOLOv3~\cite{redmon2018yolov3} can be used as \emph{sensors} in SLAM, and that their observations -- the bounding boxes around objects -- can directly constrain dual quadric parameters via our novel \emph{geometric} error formulation. To incorporate quadrics into SLAM, we derive a factor graph-based SLAM formulation that jointly estimates the dual quadric and robot pose parameters assuming solved data association. We provide a large-scale evaluation using 250 indoor trajectories through a high-fidelity simulation environment in combination with real world experiments on the TUM RGB-D~\cite{sturm12iros} dataset to show how object detections and dual quadric parametrization aid the SLAM solution.

Previous work \cite{Rubino17} utilized dual quadrics as a parametrization for landmark \emph{mapping} %
only, was limited to an orthographic camera \cite{Crocco16}, or used an \emph{algebraic} error that proved to be invalid when landmarks are only partially visible \cite{sunderhauf2017dual}. In this new work we perform full SLAM, i.e. Simultaneous Localization \emph{And} Mapping, with a general \emph{perspective} camera and a more robust \emph{geometric} error.
Furthermore, previous work \cite{Rubino17, Crocco16} required ellipse fitting as a pre-processing step: here we show that dual quadrics can be estimated in SLAM directly from bounding boxes.

\section{Related Work}
In the following section we discuss the use of semantically meaningful landmark representations in state-of-the-art mapping systems and detail existing literature that utilizes quadric surfaces as object representations.

\subsection{Maps and Landmark Representations in SLAM}
Most current SLAM systems represent the environment as a collection of distinct geometric points that are used as landmarks. ORB-SLAM~\cite{Mur15, mur2017orb} is one of the most prominent recent examples for such a point-based visual SLAM system. Even \emph{direct} visual SLAM approaches~\cite{Engel14, Whelan15} produce point cloud maps, albeit much denser  than previous approaches. Other authors explored the utility of higher order geometric features such as line segments \cite{Lemaire07b} or planes \cite{Kaess15}.

A commonality of all those geometry-based SLAM systems is that their maps carry \emph{geometric} but no immediate \emph{semantic} information. 
An exception is the influential work by Salas-Moreno et al.~\cite{Salas13}. This work proposed an object oriented SLAM system by using real-world \emph{objects} such as chairs and tables as landmarks instead of geometric primitives. \cite{Salas13} detected these objects in RGB-D data by matching 3D models of known object instances. In contrast to \cite{Salas13}, the approach presented in this paper does not require a-priori known object CAD models, but instead uses the YOLOv3 visual object detection system.

SemanticFusion~\cite{McCormac16} recently demonstrated how a dense 3D reconstruction obtained by SLAM can be enriched with semantic information. This work, and other similar papers such as \cite{Pham15}, add semantics to the map \emph{after} it has been created and do not fully leverage the relationship between geometry and semantics. The maps are not object-centric, but rather dense point clouds, where every point carries a semantic label, or a distribution over labels. In contrast, our approach uses objects as landmarks inside the SLAM system and the resulting map consists of objects encoded as quadrics.

\subsection{Dual Quadrics as Landmark Representations}
The connection between object detections and dual quadrics was recently investigated by \cite{Crocco16} and~\cite{Rubino17}.
Crocco et al.~\cite{Crocco16} presented an approach for estimating dual quadric parameters from object detections in closed form. Their method however is limited to orthographic cameras, while our approach works with perspective cameras, and is therefore more general and applicable to robotics scenarios. Furthermore, \cite{Crocco16} requires an ellipse-fitting step around each detected object. In contrast, our method can estimate camera pose and quadric parameters directly from the bounding boxes produced by typical object detection approaches such as \cite{redmon2018yolov3, Liu16, Ren15}.

As an extention of \cite{Crocco16}, Rubino et al.~\cite{Rubino17} described a closed-form approach to recover dual quadric parameters from object detections in multiple views. Their method can handle perspective cameras, but does not solve for camera pose parameters. It therefore performs only landmark \emph{mapping} given known camera poses. In contrast, our approach performs full Simultaneous Localization And Mapping, i.e. solving for camera pose, landmark pose and shape parameters simultaneously. Similar to \cite{Crocco16}, \cite{Rubino17} also requires fitting ellipses to bounding box detections first.

We explored initial ideas of using dual quadrics as landmarks in factor-graph SLAM in \cite{sunderhauf2017dual}. This unpublished preliminary work proposed an algebraic error formulation that proved to be not robust in situations where object landmarks are only partially visible. We overcome this problem by a novel geometric error formulation in this paper. In contrast to \cite{sunderhauf2017dual}, we constrain the quadric landmarks to be ellipsoids, initialize them correctly, and present a large-scale evaluation in a high-fidelity simulation environment and on real-world image sequences.

\section{Dual Quadrics -- Fundamental Concepts} \label{sec:objectdetection}
This section explains fundamental concepts around dual quadrics that are necessary to follow the remainder of the paper. For a more in-depth coverage we refer the reader to textbooks on projective geometry such as \cite{Hartley04}.

\subsection{Dual Quadrics}
\label{sec:quadrics}

Quadrics are surfaces in 3D space that are represented by a $4 \times 4$ symmetric matrix $\vQ$. In dual form, a quadric is defined by a set of tangential planes such that the planes form an envelope around the quadric. This dual quadric $\vQ^*$ is defined so that all planes $\vpi$ fulfill $\vpi\T\vQ^*\vpi = 0$. Examples for quadrics are bodies such as spheres, ellipsoids, hyperboloids, cones, or cylinders. 

A quadric has 9 degrees of freedom. These correspond to the ten independent elements of the symmetric matrix less one for scale. We can represent a generic dual quadric with a 10-vector $\hat{\vq} = (\hat{q}_1, ..., \hat{q}_{10})$ where each element corresponds to one of the 10 independent elements of $\vQ^*$.  

When a dual quadric is projected onto an image plane, it creates a dual \emph{conic}, following the simple rule $\vC^* = \vP\vQ^*\vP\T$. Here, $\vP = \vK[\vR | \vt]$ is the camera projection matrix that contains intrinsic and extrinsic camera parameters.
Conics are the 2D counterparts of quadrics and form shapes such as circles, ellipses, parabolas, or hyperbolas. %

\subsection{Constrained Dual Quadric Parametrization} \label{sec:quadric_parameterization}

In its general form, a dual quadric can represent both closed surfaces such as spheres and ellipsoids and non-closed surfaces such as paraboloids or hyperboloids. As only the former are meaningful representations of object landmarks, we use a constrained dual quadric representation that ensures the represented surface is an ellipsoid or sphere.

Similar to \cite{Rubino17}, we parametrize dual quadrics as: 
\begin{equation}
    \vQ^* = \vZ \> \breve{\vQ}^* \> \vZ\T
    \label{eq:quadric_from_e}
\end{equation}
where $\breve{\vQ}^*$ is an ellipsoid centred at the origin, and $\vZ$ is a homogeneous transformation that accounts for an arbitrary rotation and translation. Specifically,
\begin{equation}
    \vZ =
    \begin{pmatrix}
        \vR(\vtheta) & \vt \\
        \bf{0}\T_{3} & 1 \\
    \end{pmatrix} \:\:\text{and}\:\:
    \breve{\vQ}^* = 
    \begin{pmatrix}
        s_{1}^2 & 0 & 0 & 0 \\
        0 & s_{2}^2 & 0 & 0 \\
        0 & 0 & s_{3}^2 & 0 \\
        0 & 0 & 0 & -1 \\
    \end{pmatrix}
\end{equation}
where $\vt = (t_1, t_2, t_3)$ is the quadric centroid translation, $\vR(\vtheta)$ is a rotation matrix defined by the angles $\vtheta = (\theta_1, \theta_2, \theta_3)$, and $\vs = (s_1, s_2, s_3)$ is the shape of the quadric along the three semi-axes of the ellipsoid. 
In the following, we compactly represent a constrained dual quadric with a 9-vector $\vq = (\theta_1, \theta_2, \theta_3, \:t_1, t_2, t_3, \:s_1, s_2, s_3)\T$ and reconstruct the full dual quadric $\vQ^*$ as defined in (\ref{eq:quadric_from_e}).

\begin{figure}[t] 
    \centering
     \hspace*{-0.5cm}
     \subfloat[A naive sensor model truncates the full conic bounds (red) resulting in errors when compared against ground truth detections (green).
     \label{fig:error_naive}]{
     \includegraphics[width = 0.4\linewidth]{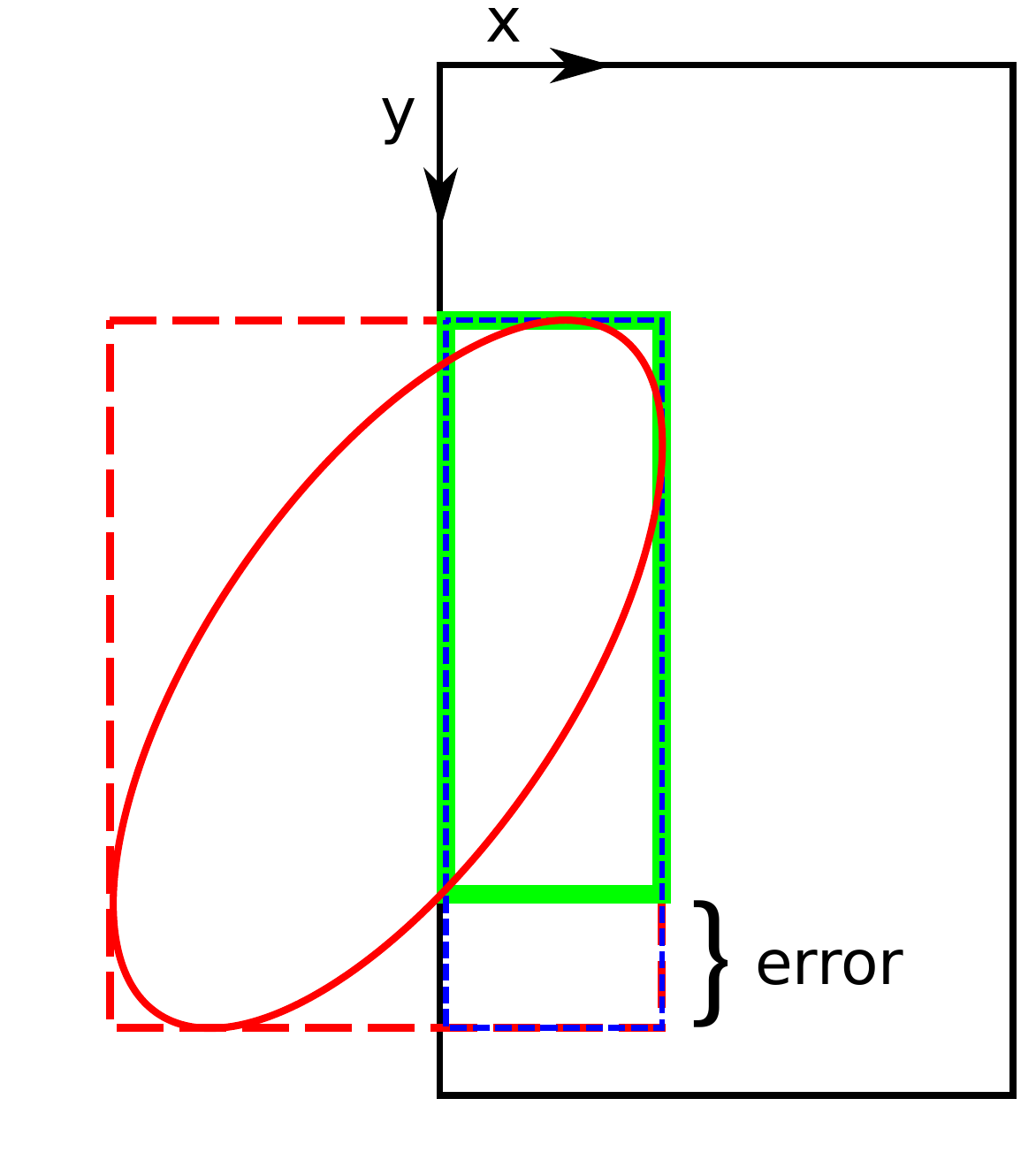}}
     \hspace*{0.25cm}
     \subfloat[Our proposed sensor model correctly predicts the object detection (green) by calculating the on-image conic bounding box (blue).
     \label{fig:error_onscreen}]{
     \includegraphics[width = 0.4\linewidth]{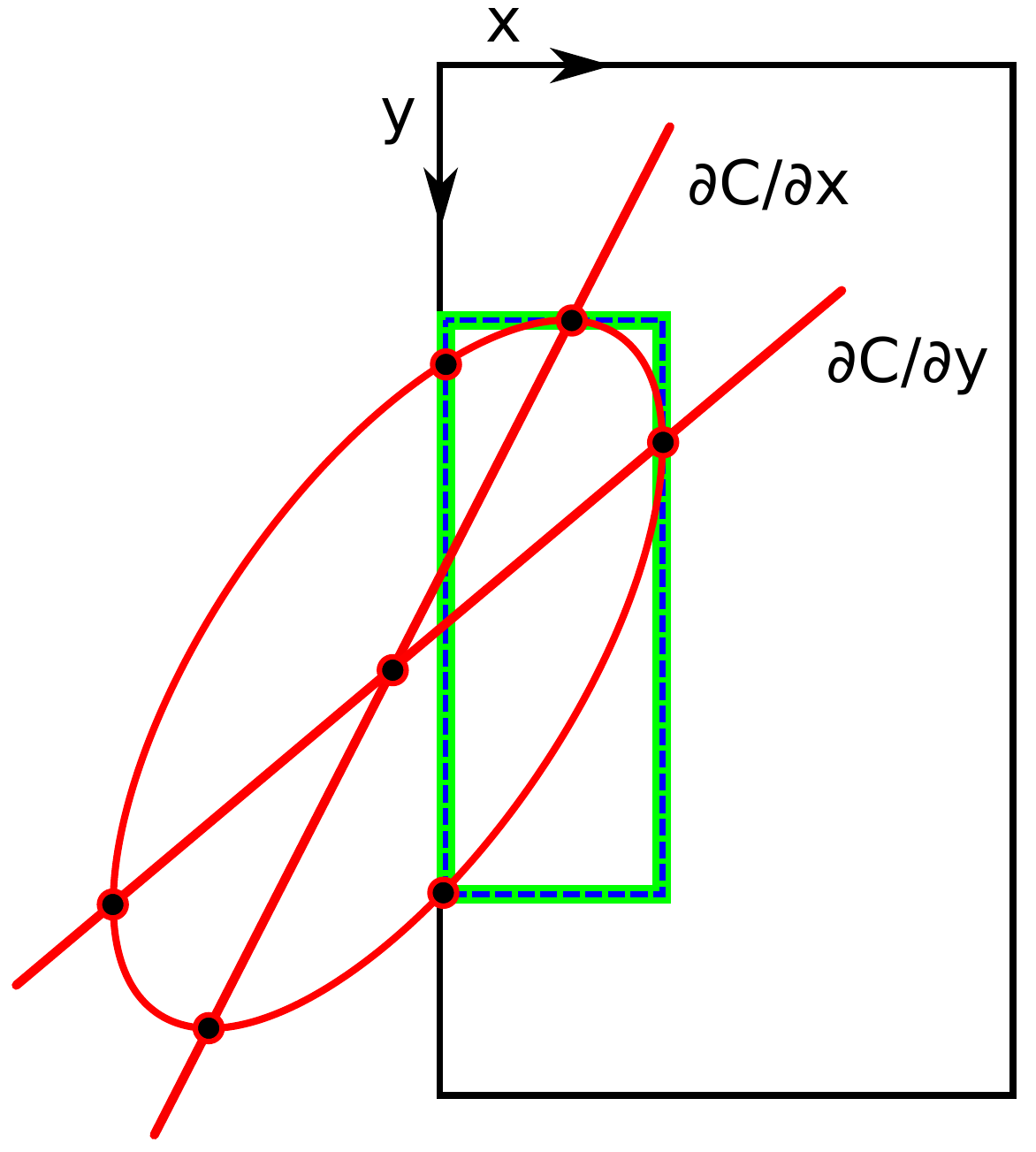}}
    \caption{Sensor models for image-based object detectors.}
\end{figure}

\section{A Sensor Model for Image-Based Object Detectors}
\label{sec:sensor_model}

\subsection{Motivation}
Our goal is to incorporate state-of-the-art object detectors such as \cite{Ren15, Liu16, He17} as a \emph{sensor} into SLAM. We therefore have to formulate a \emph{sensor model} that can predict the observations of the object detector given the estimated camera pose $\vx_i$ and the estimated map structure, i.e. quadric parameters $\vq_j$. While such sensor models are often rather simple, e.g. when using point landmarks or laser scanners and occupancy grid maps, the sensor model for an object detector is more complex.

The observations of an object detector comprise an axis-aligned bounding box constrained to the image dimensions and a discrete label distribution for each detected object. In this paper we focus on the bounding box, which can be represented as a set of four lines $\vl_k$ or a vector $\vb = (x_\text{min}, y_\text{min}, x_\text{max}, y_\text{max})$ containing the pixel coordinates of its upper-left and lower-right corner. We therefore seek a formulation for the sensor model $\vbeta(\vx_i, \vq_j) = \hat\vb_{ij}$, mapping from camera pose $\vx_i$ and quadric $\vq_j$ to predicted bounding box observation $\hat\vb_{ij}$.

This sensor model allows us to formulate a \emph{geometric error} term between the predicted and observed object detections, which is the crucial component of our overall SLAM system as explained in Section~\ref{sec:quadric_SLAM}.

\subsection{Deriving the Object Detection Sensor Model $\vbeta$}
Our derivation of $\vbeta(\vx_i, \vq_j) = \hat\vb_{ij}$ starts with projecting the estimated quadric parametrized by $\vq_j$ into the image using the camera pose $\vx_i$ according to $\vC_{ij}^* = \vP_{i}\vQ_{(\vq_{j})}^*\vP_{i}\T$ with $\vP = \vK[\vR | \vt]$ comprising the intrinsic ($\vK$) and pose parameters of the camera. Given the dual conic $\vC^*$, we obtain its primal form $\vC$ by taking the adjugate. 

A naive sensor model would simply calculate the enclosing bounding box of the conic $\vC$ and truncate this box to fit the image. However, as illustrated in Figure~\ref{fig:error_naive}, this can introduce significant errors when the conic's extrema lie outside of the image boundaries.

An accurate sensor model requires knowledge of the intersection points between conic and image borders.
The correct prediction of the object detector's bounding box therefore is the minimal axis aligned rectangle that envelopes all of the conic contained within the image dimensions. We will explain the correct method of calculating this conic bounding box, denoted \texttt{BBox($\vC$}), below. 
The overall sensor model is then defined as 
\begin{equation}
\vbeta(\vx_i, \vq_j) = \texttt{BBox}\left(\operatorname{adjugate}(\vP\vQ_{(\vq_{j})}^*\vP\T)\right) = \hat\vb_{ij}    
\end{equation}

\subsection{Calculating the On-Image Conic Bounding Box}
We can calculate the correct on-image conic bounding box by the following algorithm we denote \texttt{BBox($\vC$}):
\begin{enumerate}
    \item Find the four extrema points of the conic $\vC$, i.e. the points $\{\vp_1, ..., \vp_{4}\}$ on the conic that maximise or minimise the $x$ or $y$ component respectively.
    \item Find the up to 8 points $\{\vp_5, ..., \vp_{12}\}$ where the conic intersects the image boundaries.
    \item Remove all non-real points and all points outside the image boundaries from the set $\cP=\{\vp_1, ..., \vp_{12}\}$.
    \item Find and return the maximum and minimum $x$ and $y$ coordinate components among the remaining points.
\end{enumerate}

The function \texttt{BBox($\vC$)} therefore executes all the above steps and returns a vector
\begin{equation}
    \hat\vb = (
    \underset{x}{\text{min}}\;\hat{\cP}, 
    \underset{y}{\text{min}}\;\hat{\cP}, 
    \underset{x}{\text{max}}\;\hat{\cP}, 
    \underset{y}{\text{max}}\;\hat{\cP})    
\end{equation}
that correctly describes a bounding box that envelopes the portion of the conic $\vC$ that would be visible in the image.

\section{SLAM with Dual Quadric Landmark Representations}
\label{sec:quadric_SLAM}

\subsection{General Problem Setup}
We will set up a SLAM problem where we have odometry measurements $\vu_i$ between two successive poses $\vx_i$ and $\vx_{i+1}$, so that $\vx_{i+1} = f(\vx_i, \vu_i) + \vw_i$.
Here $f$ is a usually nonlinear function that implements the motion model of
the robot and the $\vx_i$ and $\vx_{i+1}$ are the unknown robot poses. $\vw_i$ are zero-mean Gaussian error terms with covariances $\Sigma_i$.
The source of the odometry measurements $\vu_i$ is not of concern for the following discussion, and various sources such as wheel odometers or visual odometry are possible.

We furthermore observe a set of detections $B = \{\vb_{ij}\}$. We use this notation to indicate a bounding box around an object $j$ being observed from pose $\vx_i$. Notice that we assume the problem of \emph{data association} is solved, i.e. we can identify which physical object $j$ the detection originates from\footnote{For a discussion of SLAM methods robust to data association errors see the relevant literature such as \cite{Suenderhauf12e, Agarwal13}. The methods discussed for pose graph SLAM can be adopted to the landmark SLAM considered here.}.

\subsection{Building and Solving a Factor Graph Representation}
The conditional probability distribution over all robot poses $X=\{\vx_i\}$, and landmarks $Q=\{\vq_j\}$, given the observations $U=\{\vu_i\}$, and $B=\{\vb_{ij}\}$ can be factored as
\begin{equation}
   P(X,Q|U,B) \propto \underbrace{\prod_i P(\vect{x}_{i+1} | \vect{x}_i,
   \vect{u}_{i})}_\text{Odometry Factors}
   \cdot
   \underbrace{\prod_{ij} P(\vq_j | \vect{x}_i, \vect{b}_{ij})}_\text{Landmark Factors}
  \label{eq:SLAM:posegraph:probability}
\end{equation}
This factored distribution can be conveniently modelled as a factor graph \cite{Kschischang01}. 

Given the sets of observations $U$, $B$, we seek the \emph{optimal}, i.e. maximum a posteriori (MAP) configuration of robot poses and dual quadrics, $X^*$, $Q^*$ to solve the landmark SLAM problem represented by the factor graph. This MAP variable configuration is equal to the mode of the joint probability distribution $P(X,Q)$. In simpler words, the MAP solution is the point where that distribution has its maximum. 

The odometry factors $P(\vx_{i+1} | \vx_i, \vu_i)$ are typically assumed to be Gaussian, i.e. $\vx_{i+1} \sim \cN(f(\vx_i, \vu_i), \vSigma_i)$, where $f$ is the robot's motion model. To integrate the landmark factors into a Gaussian factor graph, we apply Bayes rule:

\begin{equation}
    P(\vq_j | \vect{x}_i, \vect{b}_{ij}) = \frac{P(\vb_{ij}|\vq_j, \vx_i)\cdot P(\vq_j|\vx_i)}{P(\vb_{ij}|\vx_i)} 
\end{equation}
Since we are performing MAP estimation, we can ignore the denominator which essentially serves as a normalizer. Furthermore, assuming a uniform prior $P(\vq_j|\vx_i)$, we see that maximizing the posterior $P(\vq_j | \vect{x}_i, \vect{b}_{ij})$ is equivalent to maximizing the likelihood term $P(\vb_{ij}|\vq_j, \vx_i)$. This likelihood can be modelled as a Gaussian $\cN(\vbeta_{(\vx_i,\vq_j)}, \vLambda_{ij})$, where $\vbeta$ is the sensor model defined in Section \ref{sec:sensor_model}, and $\vLambda$ is the Covariance matrix capturing the spatial uncertainty (in image space) of the observed object detections.

The optimal variable configuration ${X^*, Q^*}$ can now be determined by maximizing the joint probability \eqref{eq:SLAM:posegraph:probability}. We formulate this as a nonlinear least squares problem by taking the negative log and factoring the joint probability:
\begin{align}
  X^*, Q^* &= 
      \argmin_{X,Q} -\log P(X,Q|U,B) \nonumber\\
      &=\argmin_{X,Q} 
      \underbrace{\sum_i \|f(\vect{x}_i, \vect{u}_i) \ominus \vect{x}_{i+1}\|^2_{\Sigma_{i}}}_\text{Odometry Factors}  \nonumber\\
      & \hspace{1.2cm}+ \underbrace{\sum_{ij}  \|\vb_{ij} - \vbeta_{(\vx_i,\vq_j)}\|^2_{\Lambda_{ij}}}_\text{Quadric Landmark Factors}
      \label{eq:SLAM_lsq}
\end{align}

Here $\|a-b\|^2_\Sigma$ denotes the squared Mahalanobis distance with covariance $\Sigma$. We use the $\ominus$ operator in the odometry factor to denote the difference operation is carried out in SE(3) rather than in Euclidean space.

Nonlinear least-squares problems such as (\ref{eq:SLAM_lsq}) can be solved iteratively  using methods like Levenberg-Marquardt or Gauss-Newton. Solvers that exploit the sparse structure of the factorisation can solve typical problems with thousands of variables very efficiently.

\subsection{The Geometric Error Term}
The error term $\|\vb_{ij} - \vbeta_{(\vx_i, \vq_j)}\|^2_{\Lambda_{ij}}$ that constitutes the quadric landmark factors in \eqref{eq:SLAM_lsq} is a \emph{geometric} error, since $\vb$ and $\vbeta$ are vectors containing pixel coordinates. In contrast to the \emph{algebraic} error proposed in previous work \cite{Crocco16, Rubino17, sunderhauf2017dual}, we found our geometric error formulation is well-defined even when the observed object is only partially visible in the image (see Figure~\ref{fig:error_onscreen}). Such situations result in truncated bounding box observations that invalidate the algebraic error formulation and shrink the estimated quadric. 

Using a geometrically meaningful error will also allow us to conveniently propagate the spatial uncertainty of the object detector (e.g. as proposed by~\cite{Miller18}) into the SLAM system via the covariance matrices $\vLambda_{ij}$ in future work.

\subsection{Variable Initialization}
\label{sec:initialization}
All variable parameters $\vx_i$ and $\vq_j$ must be initialized in order for the incremental solvers to work. While the robot poses $\vx_i$ can be initialized to an initial guess obtained from the raw odometry measurements $\vu_i$, initializing the dual quadric landmarks $\vq_j$ requires more consideration. 

It is possible to initialize $\vq_j$ with the least squares fit to its defining equation:
\begin{equation}
  \label{eq:q_init_1}
  \vpi_{ijk}\T\vQ^*_{(\hat{\vq_j})}\vpi_{ijk} = 0  
\end{equation}
where $\hat{\vq_j}$ is the vector form of a general dual quadric defined in Section~\ref{sec:quadrics}, not to be confused with the parametrized quadric vector $\vq_j$ presented in Section~\ref{sec:quadric_parameterization}. 

We can form the homogeneous vectors defining the planes $\vpi_{ijk}$ using the landmark bounding box observations $\vb_{ij}$ and resulting lines $\vl_{ijk}$ by projecting them according to $\vpi_{ijk} = \vP_i\T\vl_{ijk}$. Here the camera matrix $\vP_i$ is formed using the initial camera pose estimates $\vx_i$ obtained from the odometry measurements. 
Exploiting the fact that $\vQ^*_{(\hat{\vq_j})}$ is symmetric, we can rewrite (\ref{eq:q_init_1}) for a specific $\vpi_{ijk}$ as:
\begin{align}
  \label{eq:q_init_2}
  &(\pi_1^2,\; 2\pi_1\pi_2,\; 2\pi_1\pi_3,\; 2\pi_1\pi_4,\; \pi_2^2,\; 2\pi_2\pi_3,\; ...,\nonumber\\
  &\hspace{0.8cm} 2\pi_2\pi_4,\; \pi_3^2,\; 2\pi_3,\; \pi_4^2) \cdot (\hat{q}_1, \hat{q}_2, ..., \hat{q}_{10})\T = 0 
\end{align}
By collecting all these equations that originate from multiple views $i$ and planes $k$, we obtain a linear system of the form $\vA_j \hat{\vq_j} = 0$ with $\vA_j$ containing the coefficients of all $\vpi_{ijk}$ associated with observations of landmark $\hat\vq_j$ as in (\ref{eq:q_init_2}). A least squares solution $\hat\vq_j$ that minimizes $\|\vA_j \hat\vq_j\|$ can be obtained as the last column of $\vV$, where $\vA_j \hat\vq_j = \vU\vD\vV\T$ is the singular-value decomposition (SVD) of $\vA_j \hat\vq_j$.

The solution of the SVD represents a generic quadric surface and is not constrained to an ellipsoid; we therefore parametrize each landmark as defined in Section~\ref{sec:quadric_parameterization} by extracting the quadrics rotation, translation and shape. 

As in \cite{Rubino17}, we extract the shape of a quadric considering:
\begin{align}
    \begin{pmatrix}
        s_1 \\
        s_2 \\
        s_3
    \end{pmatrix}
    = \left\lvert\sqrt{-\frac{\text{det}\;\vQ}{\text{det}\;\vQ_{33}} 
    \begin{pmatrix}
        \lambda_1^{-1} \\
        \lambda_2^{-1} \\ 
        \lambda_3^{-1}
    \end{pmatrix}}\right\rvert
\end{align}
where $\vQ_{33}$ is the $3 \times 3$ upper left submatrix of the primal quadric $\vQ$, and $\lambda_1$, $\lambda_2$, and $\lambda_3$ are the eigenvalues of $\vQ$. The rotation matrix $\vR(\vtheta)$ is equal to the matrix of eigenvectors of $\vQ_{33}$. %
Finally, the translation of a dual quadric is defined by the last column of $\vQ^*$ as a homogeneous 4-vector such that $\vt = (\hat{q}_4, \hat{q}_7, \hat{q}_9) / \hat{q}_{10}$. We can then reconstruct the constrained equivelent of the estimated quadric as in Section~\ref{sec:quadric_parameterization}.

Hence, we initialize all landmarks by calculating the SVD solution of (\ref{eq:q_init_1}) over the complete set of detections for each landmark, and constrain the estimated quadrics to be ellipsoids.

\section{Experiments and Evaluation}
We evaluate the use of quadric landmarks on a number of sequences from the publicly available TUM RGB-D~\cite{sturm12iros} dataset, and compare the localization performance against state-of-the-art techniques. These real world image sequences demonstrate the effectiveness of our method under realistic conditions. We further evaluate the quality of the constructed landmarks in a high-fidelity simulation environment with ground truth 3D object shapes and positions.

\subsection{TUM RGB-D Experiments}
\label{sec:tum_experiments}
Real world datasets provide an excellent opportunity to expose quadric landmarks to an array of challenges. Image noise, motion blur, poor focus, dynamic lighting, occlusions, and moving objects are just some of the difficulties we face in the real world. We use 4 sequences from the TUM RGB-D\footnote{Depth information is used to provide a consistent scale for odometry and does not yet support the estimation of quadric parameters.} dataset (fr1\_desk, fr1\_desk2, fr2\_desk, fr3\_office) to evaluate the estimated camera trajectory under realistic conditions. In these experiments, we compare the Absolute Trajectory Error (ATE) of the estimated camera poses against ORB-SLAM2~\cite{mur2017orb}, as well as our sources of odometry measurements.

We evaluate the localization performance of our method given two sources of visual odometry. Fovis~\cite{huang2017visual} (QuadricSLAM\textsubscript{f}) and ORB-SLAM2 with loop closures disabled (QuadricSLAM\textsubscript{o}). 
Object detections have been generated using the YOLOv3~\cite{redmon2018yolov3} detector with pretrained weights, and associations between individual detections and distinct physical objects have been provided by a set of manual annotations.

We implement the SLAM problem \eqref{eq:SLAM_lsq}, coined QuadricSLAM, as a factor graph using GTSAM~\cite{gtsam}. The robot poses and dual quadrics, $X^*$ and $Q^*$, populate the latent variables of the graph, connected with odometry factors $U$, and 2D bounding box factors $B$. 
As a rough approximation for the covariance matricies of these factors, the odometry noise model is set to a standard deviation of 0.001 for both translation and rotation, and the bounding box standard deviation is approximated as the sum of width and height standard deviations (in pixels) for the entire set of bounding boxes for each object. In reality, the confidence of individual measurements varies considerably, however, more difficult to detect objects typically have a larger variance in bounding box dimensions, and so we use this as an approximation to the detection noise. %

In order to classify our quadric landmarks, we combine the classification of individual detections by averaging the full set of detection scores belonging to each object, and assigning the most likely class as the objects final classification. 

\begin{table*}[t]
\centering
\caption{
Average localization errors on TUM sequences 
} 
\label{tab:tum_results}
    \begin{tabular}{@{}lccccc@{}}
    \toprule
        Sequence & fovis & QuadricSLAM\textsubscript{f} & ORB-VO & QuadricSLAM\textsubscript{o} & ORB-SLAM2 \\ \midrule
        fr1\_desk    & 0.2589 & 0.0632 & \textbf{0.0153}    & 0.0167           & 0.0159          \\
        fr1\_desk2   & 0.1248 & 0.0662 & 0.0245             & 0.0245            & \textbf{0.0243} \\
        fr2\_desk    & 0.1029 & 0.0568 & 0.0151             & 0.0124            & \textbf{0.0087} \\
        fr3\_office  & 0.1879 & 0.0765 & 0.0142             & 0.0230            & \textbf{0.0107} \\
    \bottomrule
    \end{tabular}
\end{table*}

\begin{figure*}[t] 
\centering
\frame{
\includegraphics[width = 0.24\linewidth]{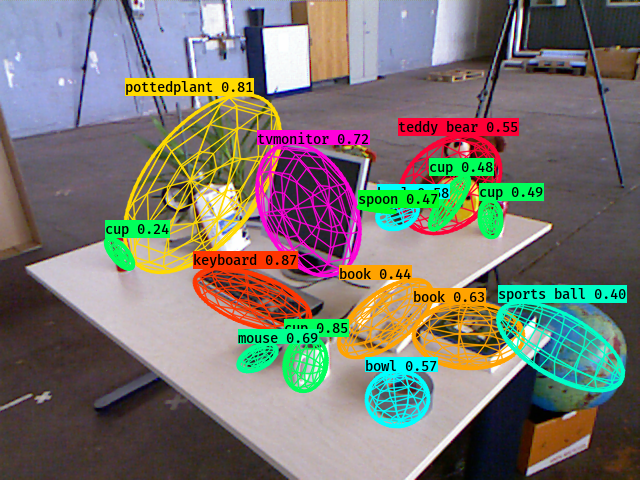}}
\frame{
\includegraphics[width = 0.24\linewidth]{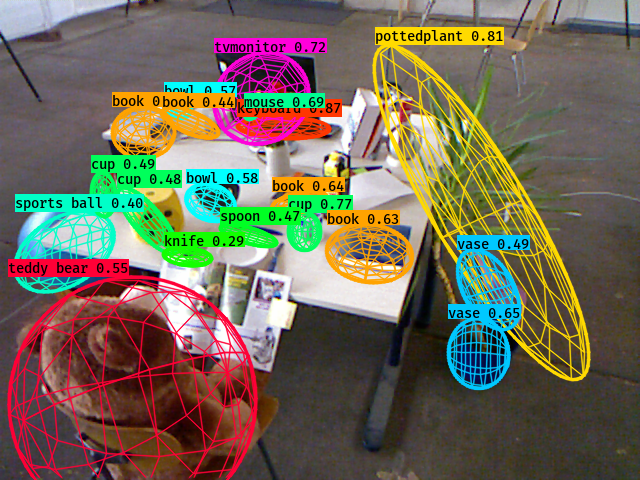}} \hspace{0.15cm}
\frame{
\includegraphics[width = 0.24\linewidth]{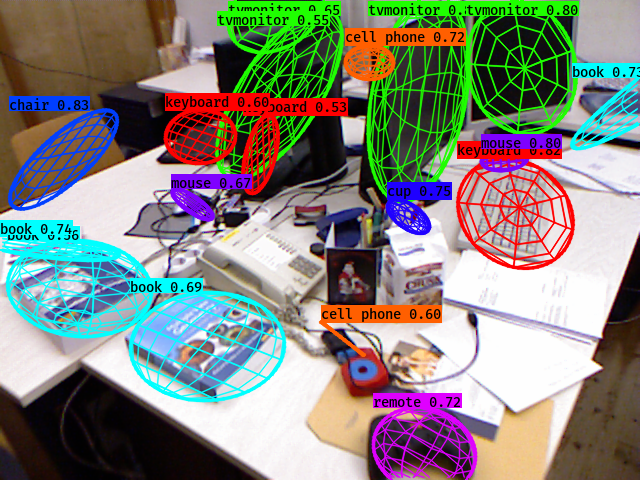}}
\frame{
\includegraphics[width = 0.24\linewidth]{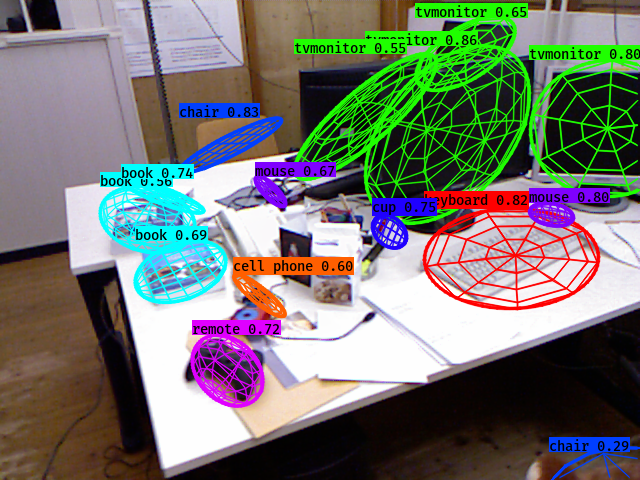}} \\ \vspace{0.1cm}
\frame{
\includegraphics[width = 0.24\linewidth]{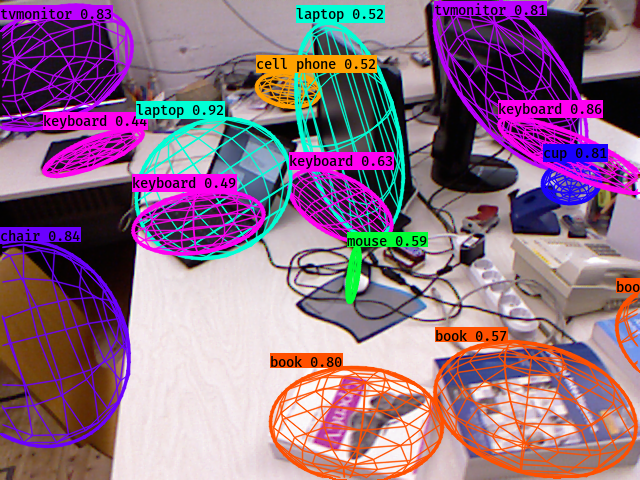}}
\frame{
\includegraphics[width = 0.24\linewidth]{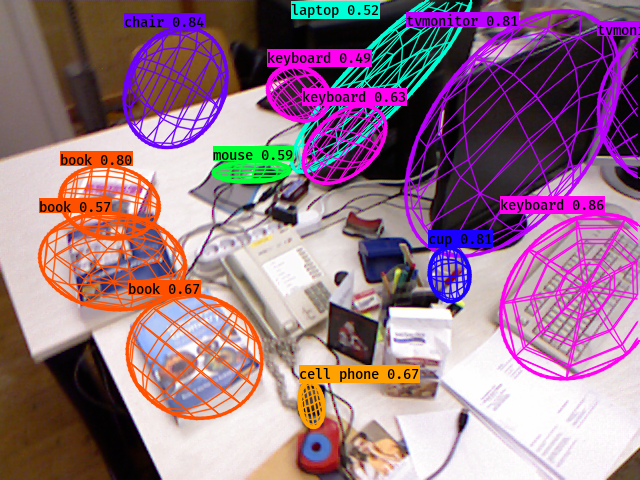}} \hspace{0.15cm}
\frame{
\includegraphics[width = 0.24\linewidth]{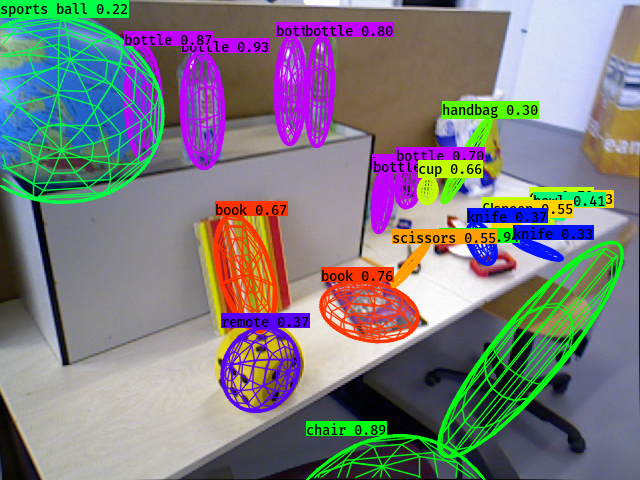}}
\frame{
\includegraphics[width = 0.24\linewidth]{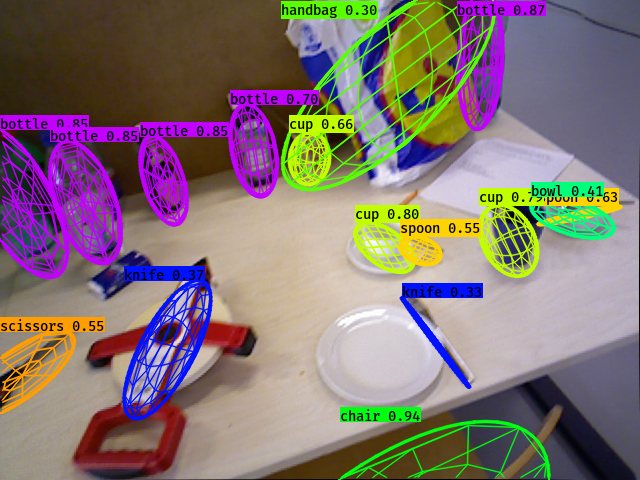}} \\ 
\caption{
The estimated landmarks for each sequence of the real-world evaluation, seen from two viewpoints each. The 3D quadrics with a visible associated object are projected into the images at each camera pose, i.e. occluded objects are not displayed. The quality of these estimations is demonstrated by the alignment of ellipsoid axes with the object boundaries. 
}
\label{fig:tum_qualitative}
\end{figure*}

\subsection{TUM RGB-D Results}
\label{sec:tum_results}

The quantitative results of our experiments on the TUM sequences can be found in Table~\ref{tab:tum_results}. These results clearly show an improvement in trajectory quality over the foviz visual odometry, but display a slight decrease in performance when compared to ORB-SLAM2. This performance loss is expected to be caused by a combination of poor constant noise estimates for the non-Gaussian bounding box measurements, and object occlusions, which create significantly smaller bounding boxes and negatively impact the estimated trajectory. 

This is supported by further experiments that demonstrated it was possible to improve the trajectory error compared to ORB-SLAM2 on a number of sequences by rejecting objects with high bounding box variance. Explicitly, by rejecting objects with a bounding box width or height standard deviation of greater than 50 pixels and 30 pixels, we were able to achieve a trajectory error of 0.0239 meters and 0.0087 meters on sequences fr1\_desk2 and fr3\_office respectively. 

We demonstrate the quality of the object-centric maps in Figure~\ref{fig:tum_qualitative}. Here, the 3 dimensional quadric surfaces have been projected into the image from the estimated camera positions. These figures highlight how accurately the ellipsoids capture the geometric shape of the objects in the scene. These images also display the predicted class label for each object.

When compared to single-view object detections, we see that these landmarks provide accurate high-level information about the 3D objects they represent, and effectively integrate the sparse and volatile single-view object detections in order to provide a consistent class label and object shape over successive frames. We have provided a supplementary video file which illustrates this point by comparing the raw detections against the generated map for a single sequence from the TUM RGB-D dataset. This will be available at http://ieeexplore.ieee.org, additional multimedia material can be found at http://semanticslam.ai. 

In some situations, we have found that the SVD solution causes some quadric landmarks to initialize behind the camera. This is possible because a quadric projected with $\vP\vQ^*\vP\T$ can project from either side of the image plane. In our experiments, this seems to occur only for landmarks that have a limited viewing angle. These objects are only observed a few times, therefore, they typically do not have a large impact on the resulting trajectory. Additionally, we found that some regions of the scene that do not contain a distinct physical object are detected enough times to construct a quadric landmark. Although these landmarks do not seem to negatively impact the trajectory, they provide misleading semantic information.

\subsection{Simulated Experiments}
\label{sec:evaluation_environment}
Simulation environments provide access to an array of rich and realistic environments and conditions, allowing us to evaluate the quality of our object landmarks against the provided ground truth object positions. With this in mind, we created a synthetic dataset using the UnrealCV plugin \cite{qiu2017unrealcv}, containing ground truth camera trajectory $\vx_i \in \text{SE(3)}$, 2D object bounding boxes $\vect{b}_{ij}$, and 3D object bounding boxes.
Trajectories were recorded over 10 scenes resulting in a total of 50 trajectories. These 50 ground truth trajectories were each corrupted with noise generated from 5 seeds for a total of 250 trials. 
The camera was simulated with a focal length of 320.0, a principal point $(C_x, C_y) = (320.0, 240.0)$, and a resolution of $640 \times 480$. These images were used to extract ground truth bounding boxes of the same form as those generated from ConvNet-based object detectors such as \cite{Ren15,Liu16}. 

Similarly to Section~\ref{sec:tum_experiments} we formulate a factor graph with robot poses and dual quadrics connected by odometry and bounding box factors. Odometry measurements for each trial are obtained by introducing zero mean Gaussian noise to the relative motion between the true camera positions. We induce an error of roughly 5\% for translation and 15\% for the rotation between global camera positions. This trajectory noise is similar to the noise found when using standard inertial navagation systems, where perturbations in the relative trajectory compound and cause the global trajectory to deviate from the true trajectory (see Figure~\ref{fig:maps}). 

Object detections were simulated by adding an additional variance of 4 pixels to the ground truth bounding box detections, and associations between 3D objects and 2D detections were provided by the simulator as we have assumed the problem of data association is solved. During the simulation experiments, we set the covariance matricies for odometry and bounding box measurements to the additional variance we have introduced. 

We evaluate the localization \emph{and} mapping performance of our method by comparing the Absolute Trajectory Error of the estimated camera trajectory against the noisy odometry measurements. Moreover, we compare the initial and estimated quadric parameters against the true 3D object bounding boxes using three metrics. We first compare the objects positions by calculating the root mean squared error (RMSE) between the estimated quadric translation and ground truth centroid. Secondly, the error in the shape of a landmark is evaluated using the Jaccard distance ($1 - \text{Intersection over Union}$) between the quadrics 3D axis aligned bounding box and the true object bounding box after centering both boxes at the origin. Finally, we evaluate the overall landmark quality using the standard Jaccard distance between both boxes.

\subsection{Simulated Results}

\begin{figure*}[t] 
\centering
\includegraphics[width = 0.24\linewidth]{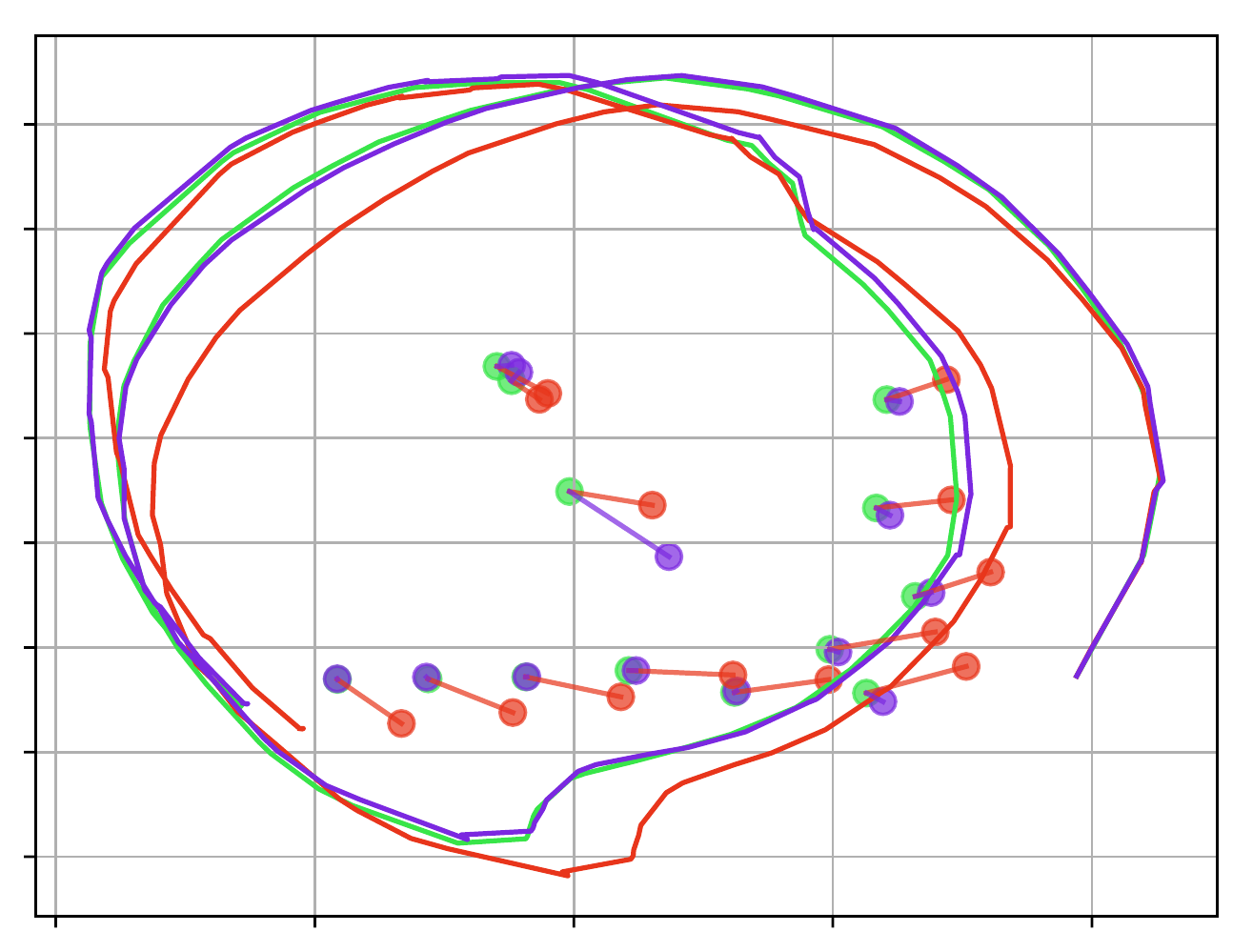}
\includegraphics[width = 0.24\linewidth]{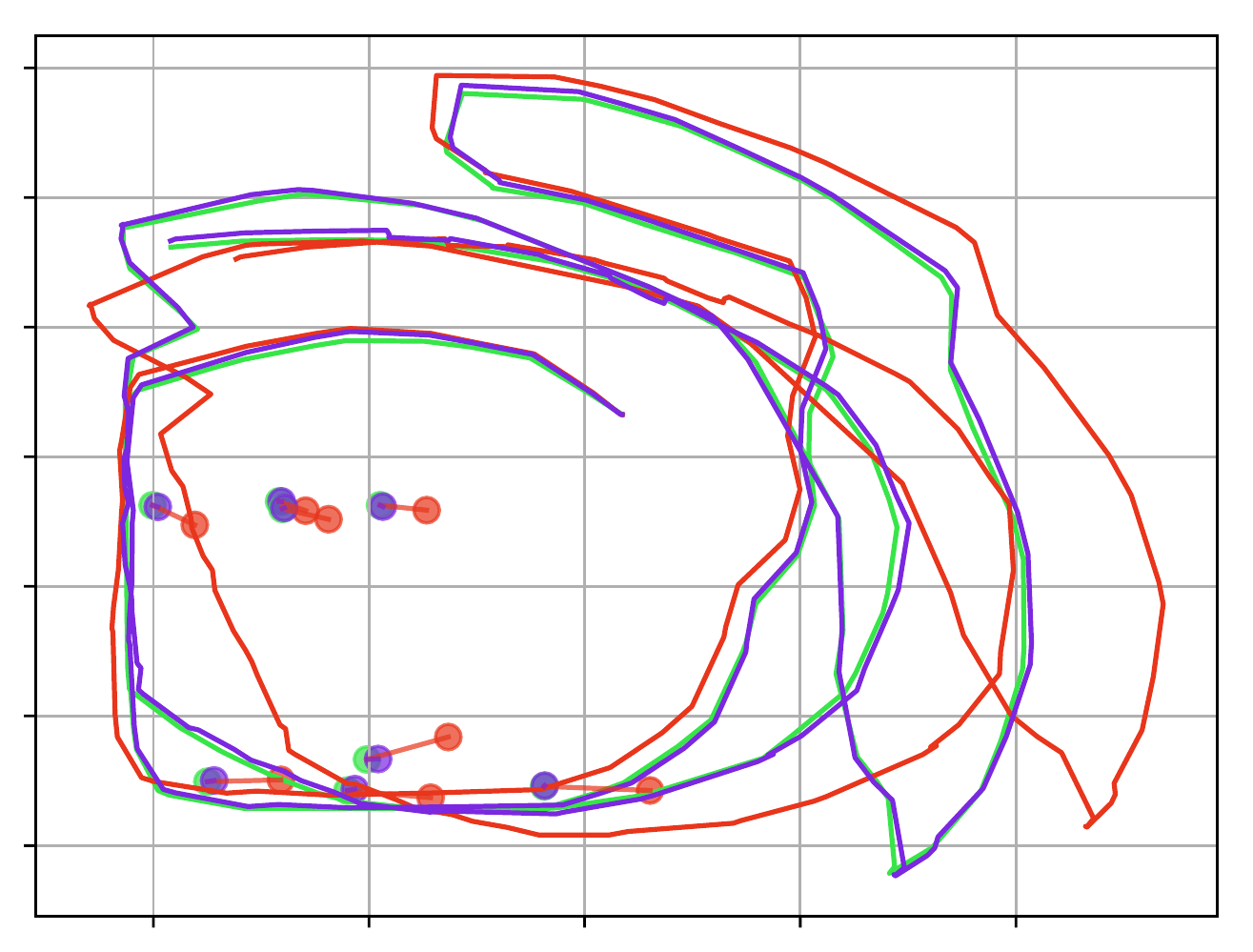}
\includegraphics[width = 0.24\linewidth]{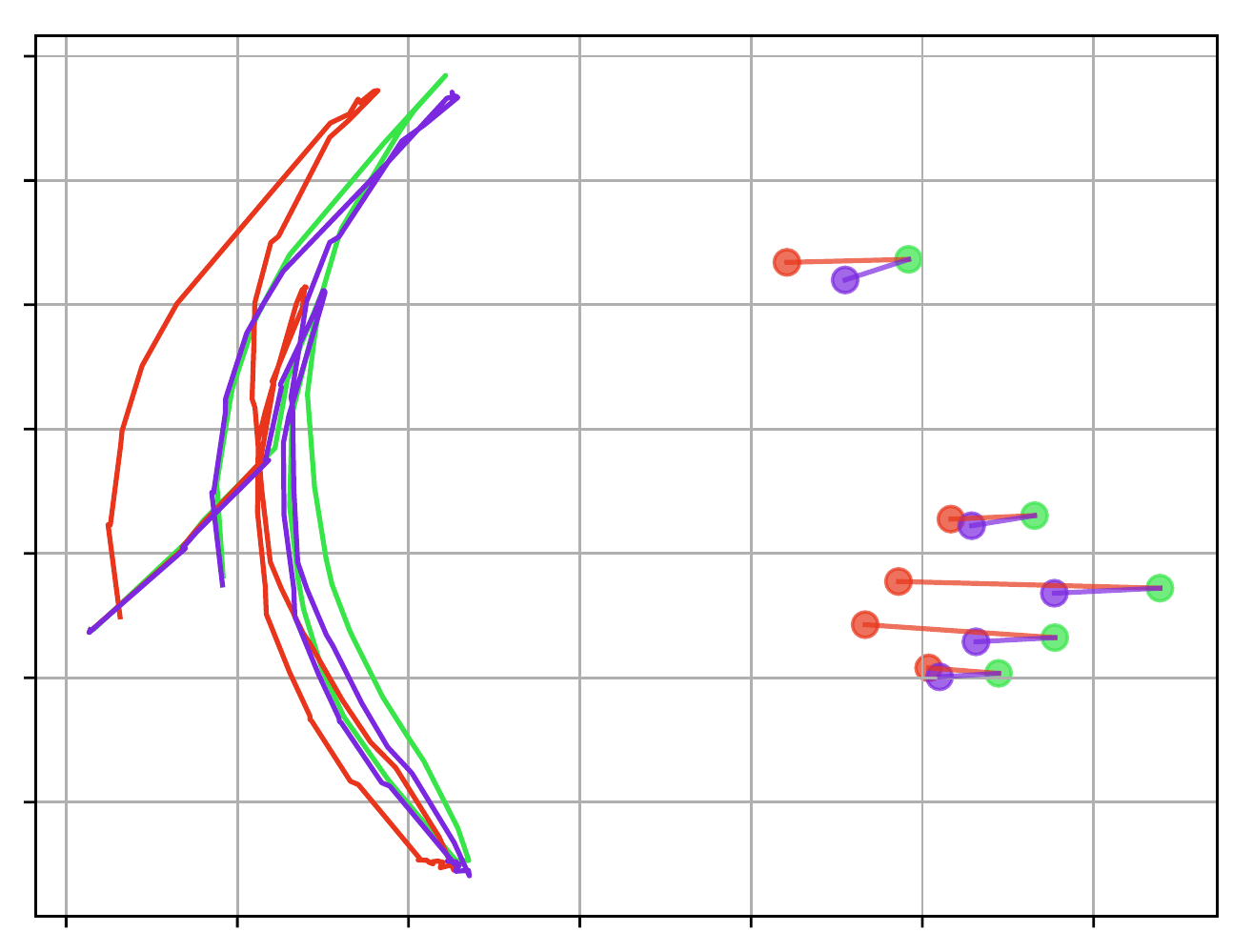} 
\includegraphics[width = 0.24\linewidth]{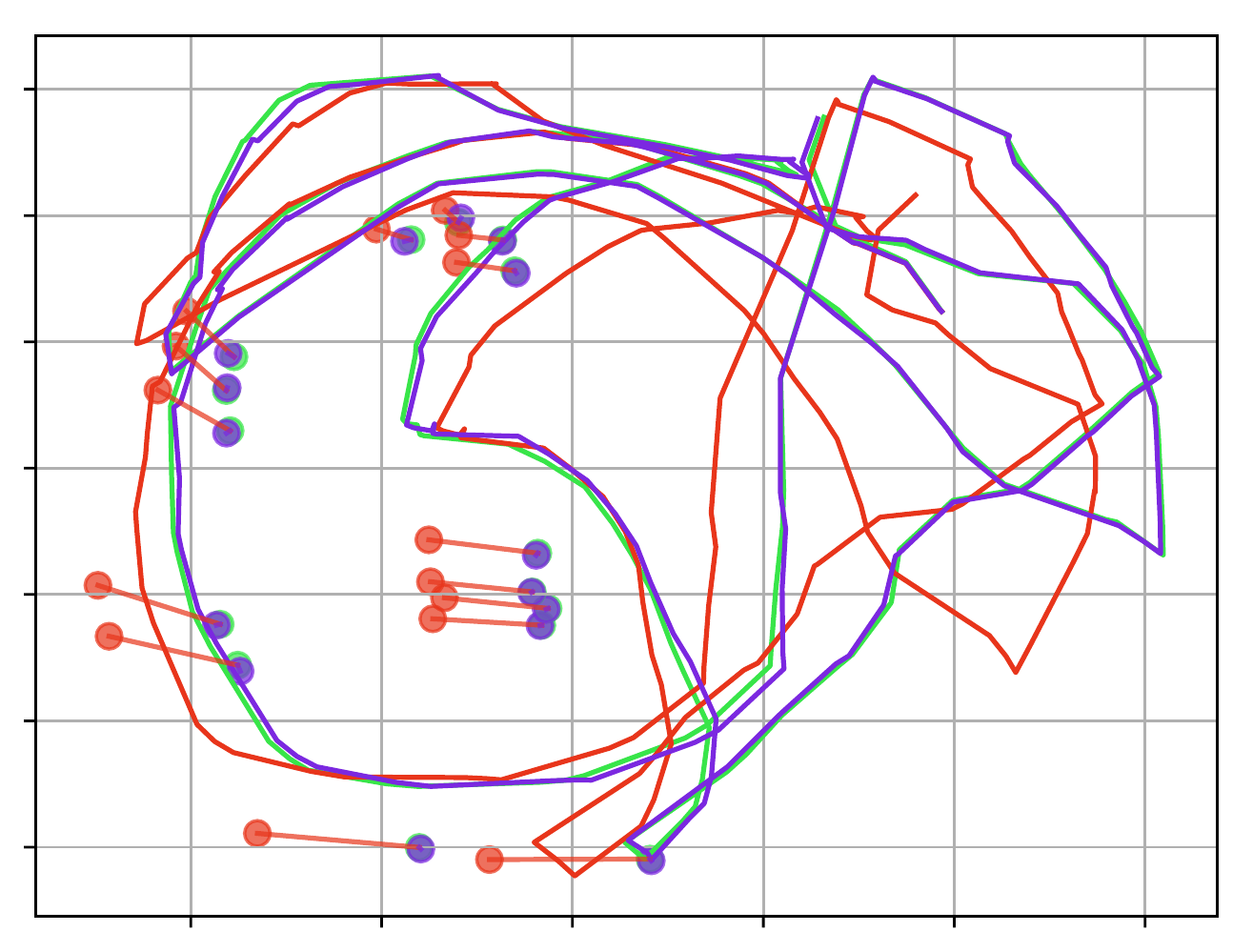}
 \caption{Example trajectories from a top-down perspective, comparing ground truth trajectory (green) with the initial odometry (red) and the trajectory estimated by our object-oriented SLAM system (blue). The plots also show the initialised (red), estimated (blue) and true (green) landmark centroids as dots, connected by red and blue lines to emphasize individual objects.}
\label{fig:maps}
\vspace{-1mm}
\end{figure*}

\begin{figure*}[t] 
\centering
\includegraphics[width = 0.24\linewidth]{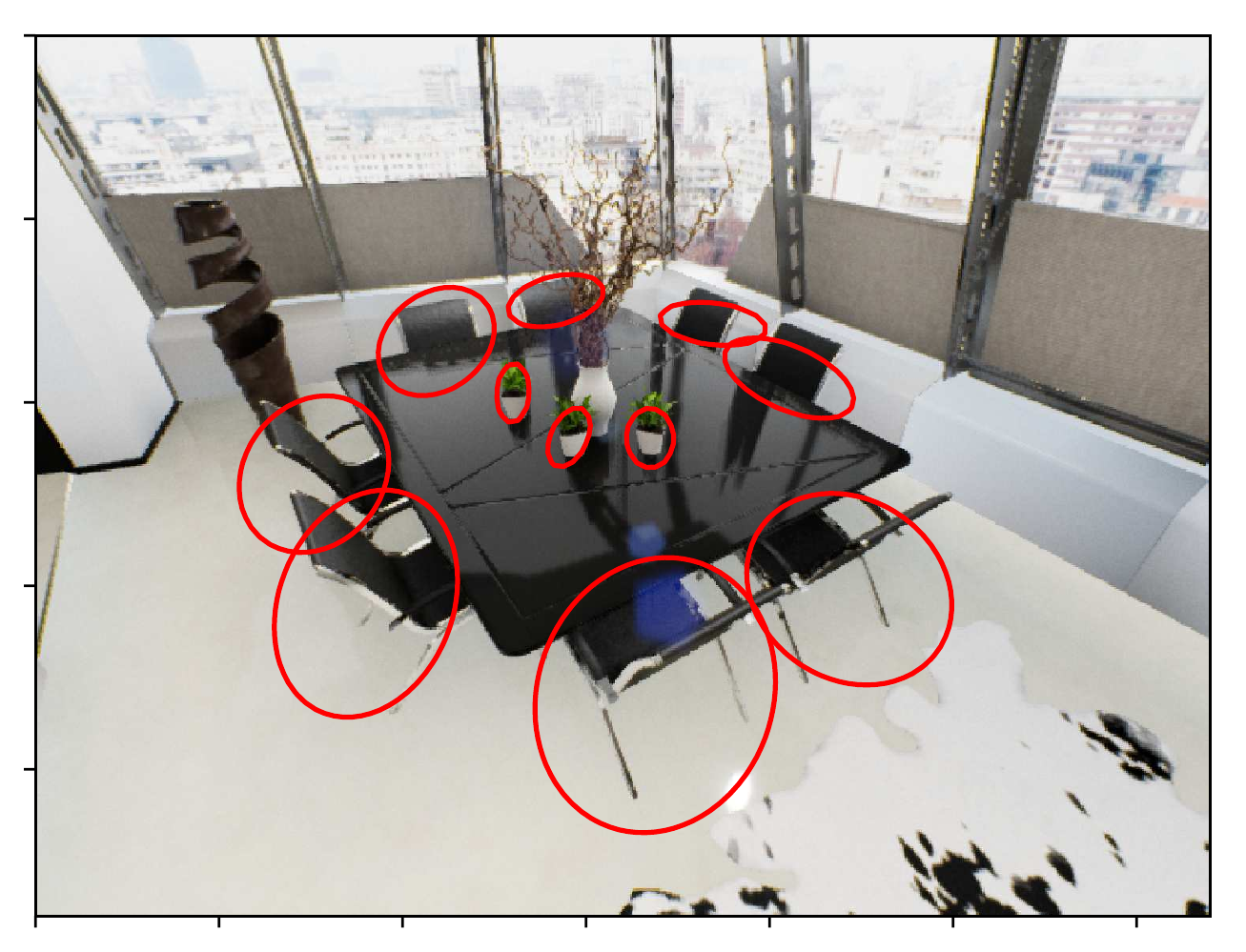}
\includegraphics[width = 0.24\linewidth]{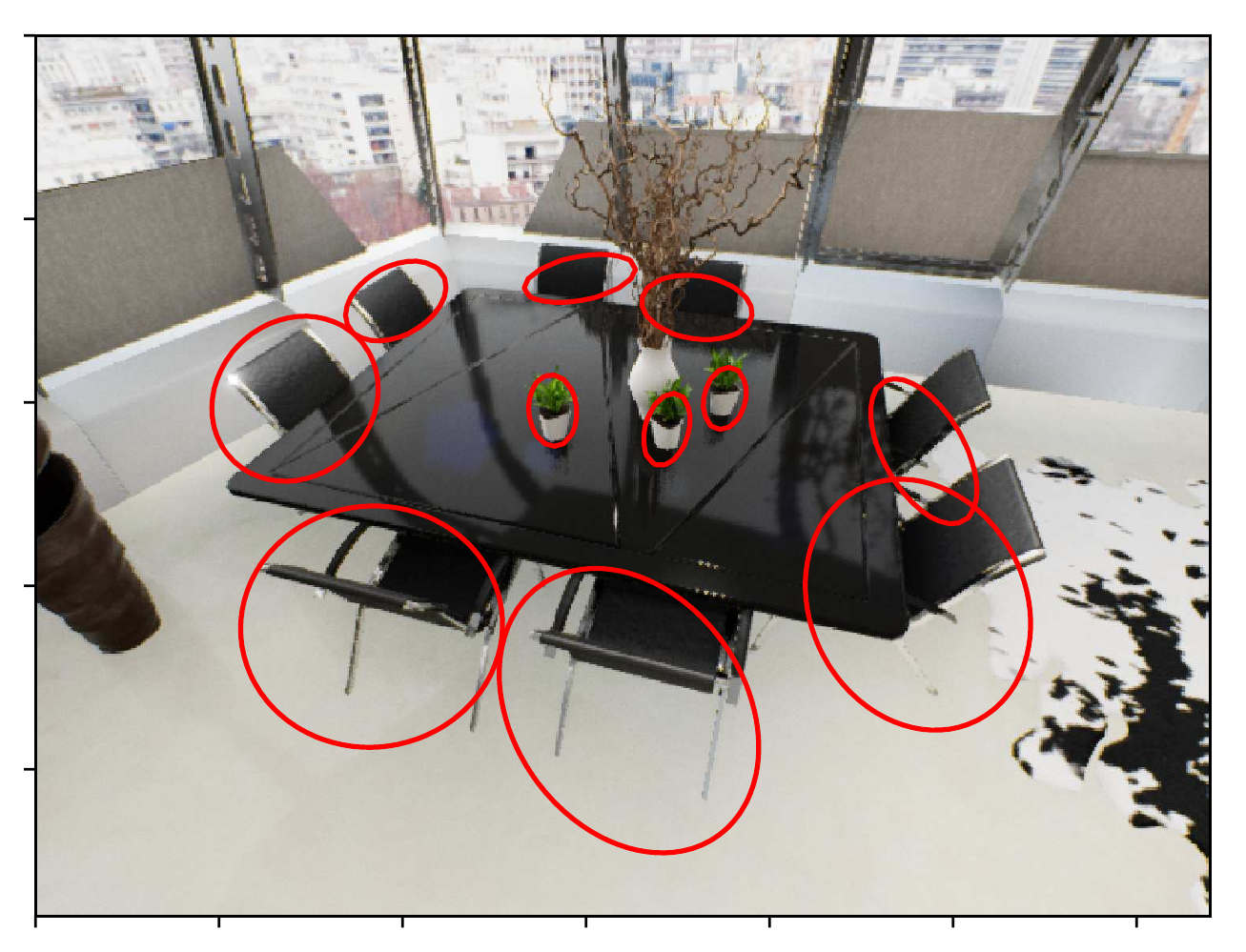} \hspace{0.15cm}
\includegraphics[width = 0.24\linewidth]{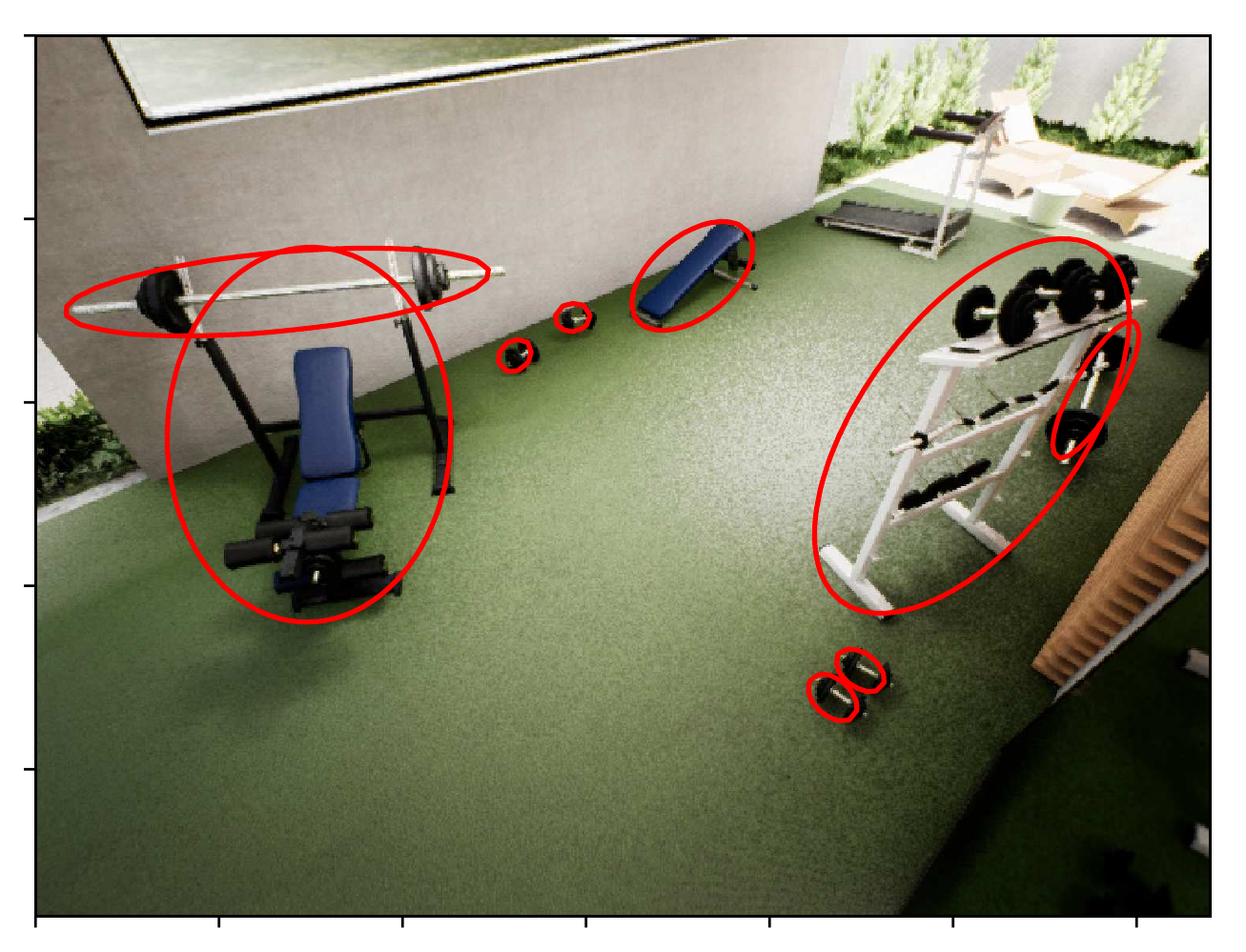}
\includegraphics[width = 0.24\linewidth]{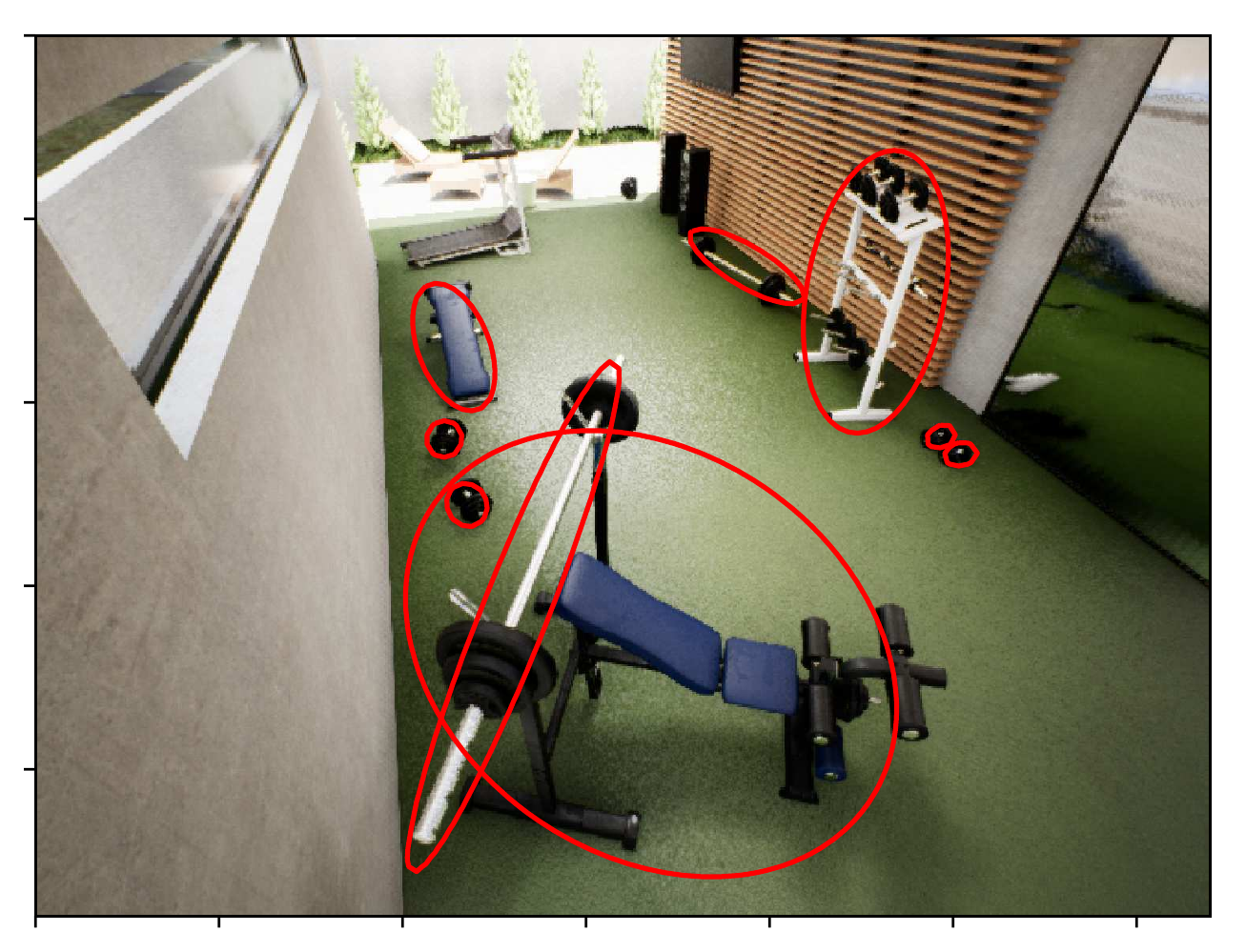} \\
\includegraphics[width = 0.24\linewidth]{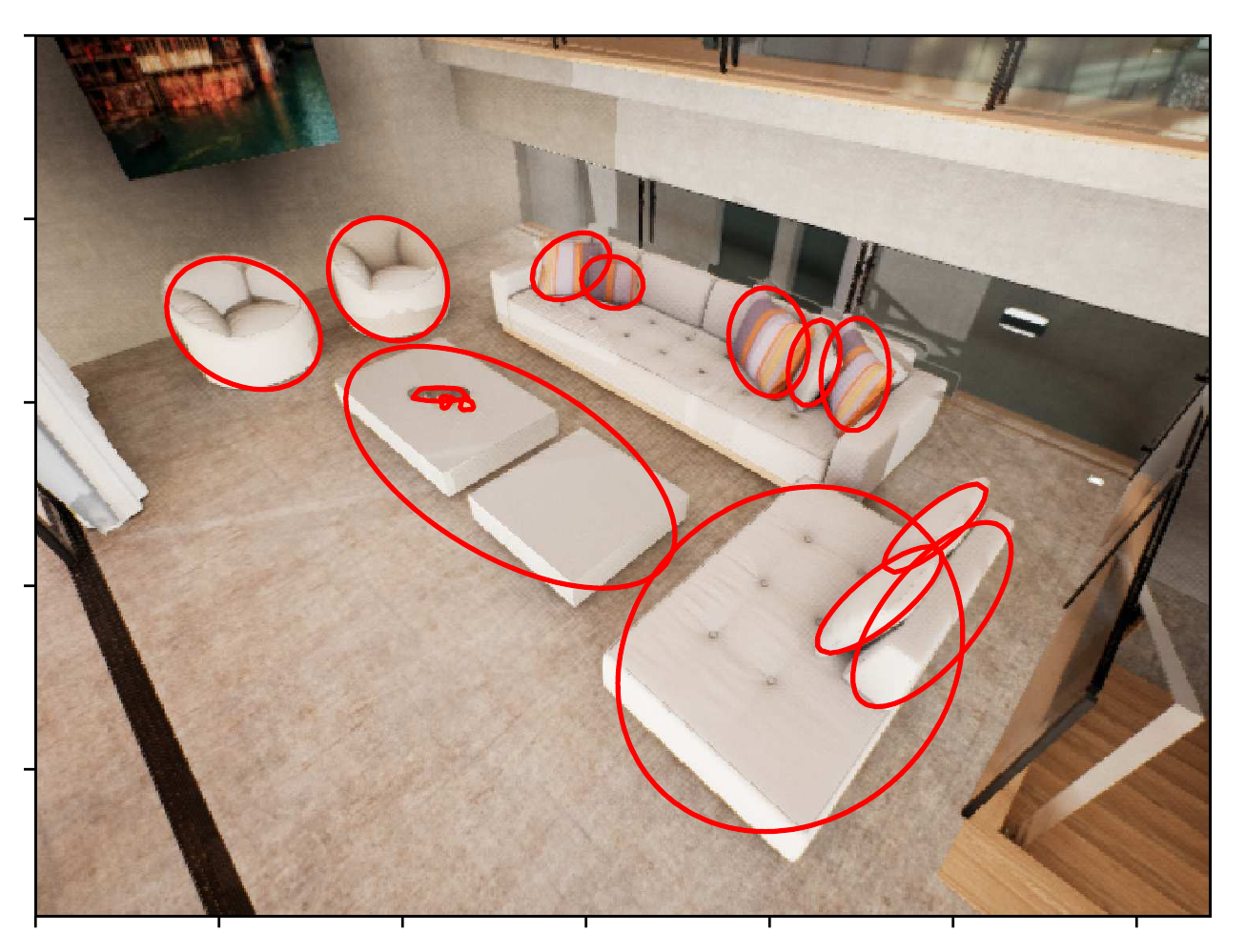}
\includegraphics[width = 0.24\linewidth]{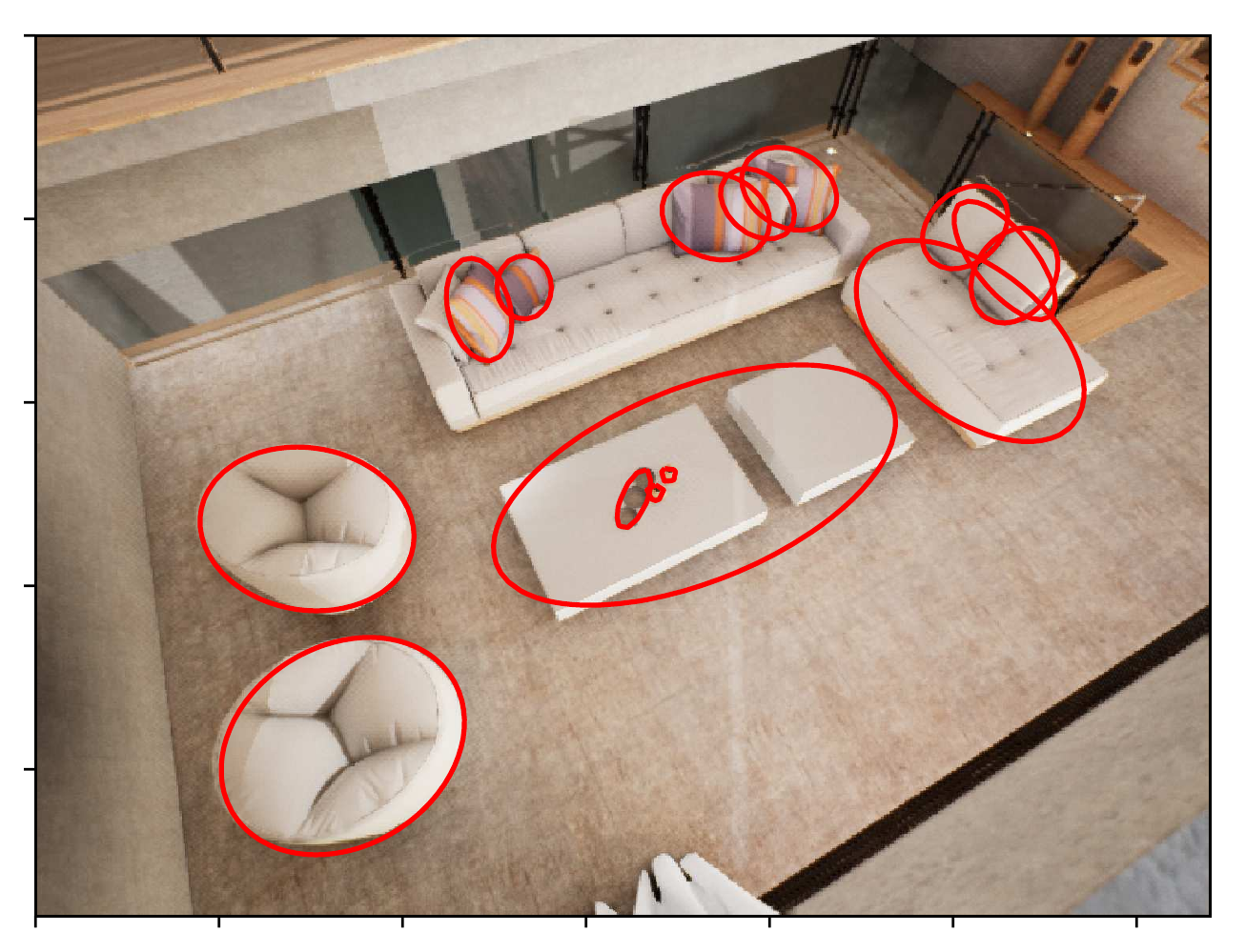} \hspace{0.15cm}
\includegraphics[width = 0.24\linewidth]{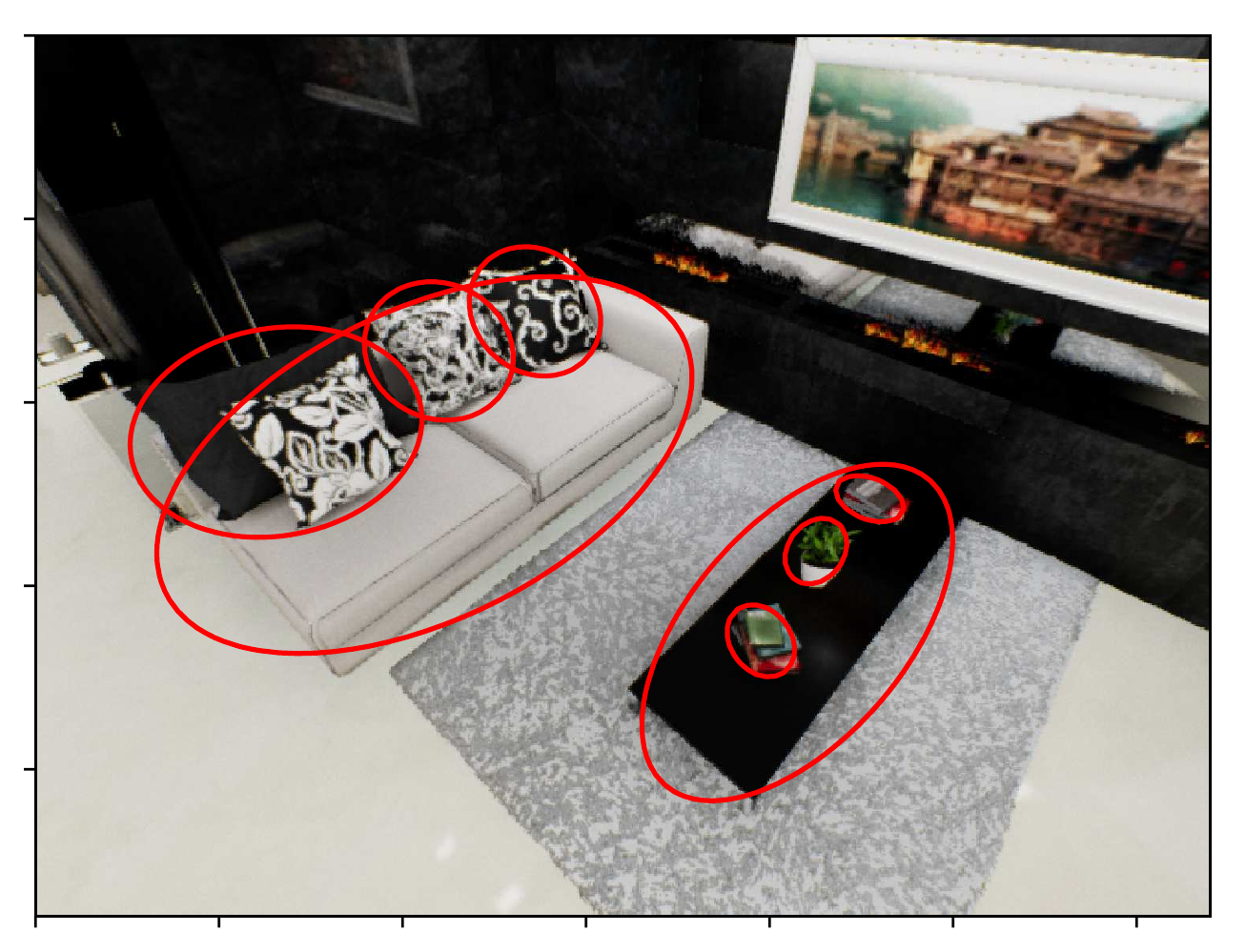}
\includegraphics[width = 0.24\linewidth]{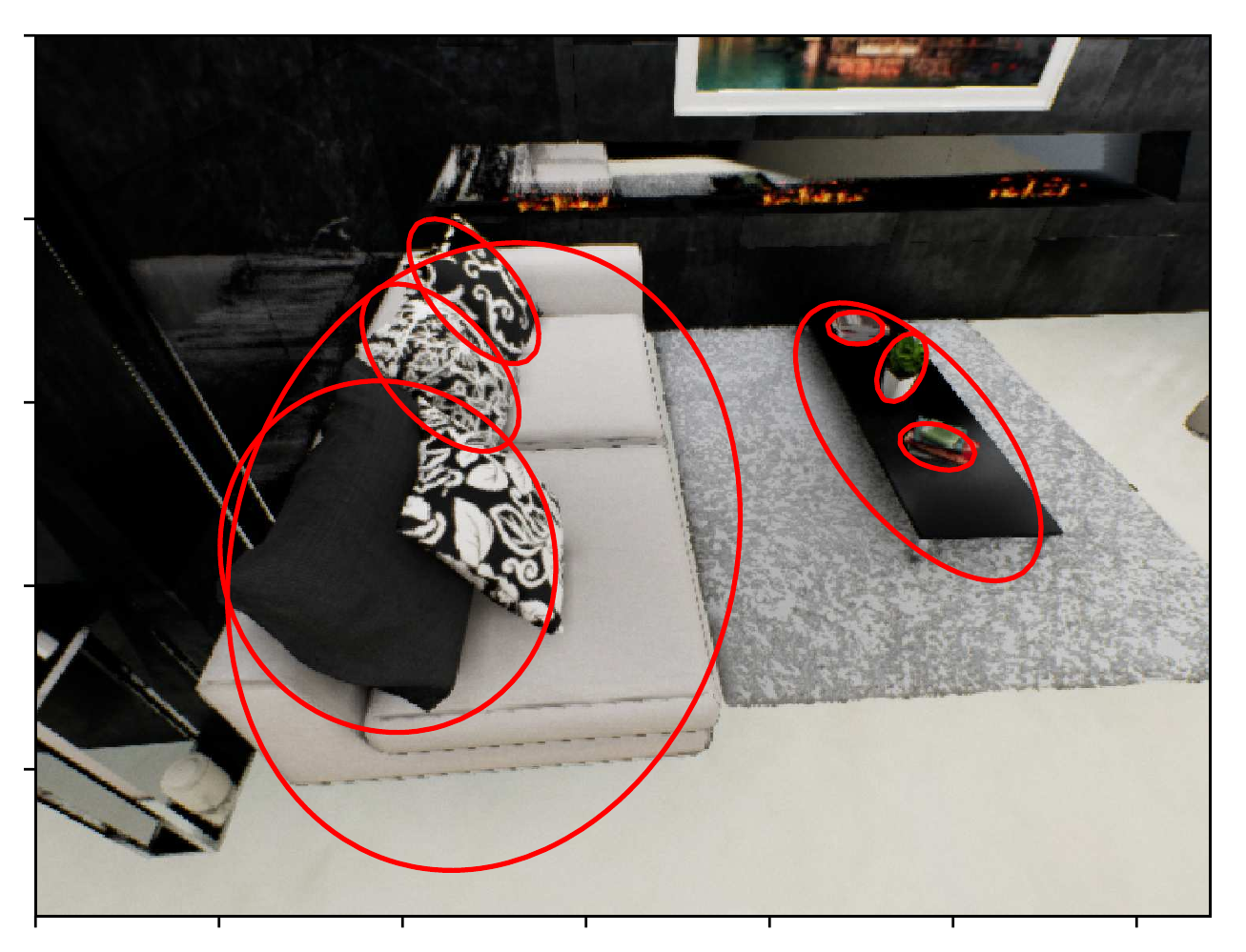} 
\caption{Estimated landmark positions and shapes (illustrated by the red ellipses) for 4 scenes of the evaluation, seen from two viewpoints each. To create these figures we projected the estimated 3D Quadrics into the images at each camera pose. The quality of these estimations is demonstrated by the alignment of major and minor ellipse axis with the object boundaries.}
\label{fig:sim_qualitative}
\vspace{-1mm}
\end{figure*}

We summarize the results of our simulated experiments in Table~\ref{tab:sim_results} and provide qualitative examples illustrating the improvement in camera trajectory and the accuracy of the estimated quadric surfaces in Figures~\ref{fig:maps} and \ref{fig:sim_qualitative} respectively.

The results show that quadric landmarks significantly improve the quality of the robot trajectory and the estimated map, providing accurate high level information about the shape and position of objects within the environment. Explicitly, the geometric error gains a 65.2\% improvement on trajectory error and a 70.4\%, 26.7\% and 30.6\% improvement on landmark position, shape and quality. 
The correcting effect of the quadric landmarks on the estimated trajectory is a result of re-observing the landmarks between frames, helping to mitigate accumulated odometry errors.

As described in Section~\ref{sec:tum_results}, the remaining discrepancies between estimated landmark parameters and ground truth objects is expected to be caused by a combination of occlusion, which encourages the shrinking of landmark surfaces, and limited viewing angles, resulting in the overestimation of landmark shapes. 

We also evaluated the performance of the standard algebraic error function utilized in previous work \cite{Crocco16, Rubino17, sunderhauf2017dual} and found that the estimated solution rarely improves on the initial map and trajectory estimate. The algebraic error improves camera trajectory and landmark quality by 0.6\% and 1.5\%, but actually negatively impacts the landmark position and shape by 2.6\% and 2.0\% respectively. This is caused by partial object visibility, exaggerated by the presence of large objects within the majority of scenes.

\begin{table*}[t]
\centering
\caption{
Average errors for localization and mapping in the simulated environment.
} 
\label{tab:sim_results}
    \begin{tabular}{@{}lcccccccc@{}}
    \toprule
     & ATE\textsubscript{trans} (m) & LM\textsubscript{trans} (m) & LM\textsubscript{shape} (\%) & LM\textsubscript{quality} (\%) \\ \midrule
        Odometry        & 0.5895 & - & - & -\\       
        SVD solution    & - & 0.5786 & 0.6073 & 0.8477\\
        QuadricSLAM     & \textbf{0.2049} & \textbf{0.1714} & \textbf{0.4450} & \textbf{0.5886}\\
        \bottomrule
    \end{tabular}
\end{table*}

\section{Conclusions and Future Work}
In conclusion, QuadricSLAM combines state-of-the-art object detection and SLAM techniques in order to simultaneously estimate the optimal camera position and a 3D landmark representation of objects within the environment. The introduction of object based landmarks, such as dual quadrics, is \emph{essential} to the development of semantically meaningful, object-oriented robotic maps. 

The results of our experiments have shown that quadric landmarks provide valuable information for correcting odometry errors, however, the most significant benefit of quadric-based object landmarks is the estimation of maps that contain objects as distinct elements. The advantages of using dual quadrics as landmark parametrizations in SLAM will only increase when incorporating higher order geometric constraints into the SLAM formulation, such as prior knowledge on how landmarks of a certain semantic type can be placed in the environment with respect to other landmarks or general structure.

Our paper has demonstrated how to use dual quadrics as landmark representations in SLAM with perspective cameras. We provided a method of parametrizing dual quadrics as closed surfaces and show how they can be directly constrained from typical object detection systems.

We developed a factor graph-based SLAM formulation that jointly estimates camera trajectory and object parameters in the presence of odometry noise, object detection noise, occlusion and partial object visibility. This has been achieved by devising a sensor model for object detectors and a geometric error that is robust to partial object observations. 

We provided an extensive evaluation of trajectory \emph{and} landmark quality in both real-world experiments and simulation, showing how our method compares to existing techniques, and demonstrating the utility of object-based landmarks for SLAM. 

Future work will investigate how we can better integrate uncertainty estimates from the object detector, explore methods of constraining the quadric initialization to be in front of the image plane, and look at using depth information to reject detections that do not correspond to a distinct physical object.  

\bibliographystyle{IEEEtran}
\bibliography{bibfile}

\end{document}